\documentclass{article}
\usepackage{ijcai18}

\usepackage{times}
\usepackage{xcolor}
\usepackage{soul}
\usepackage[utf8]{inputenc}
\usepackage[small]{caption}
\usepackage{graphicx}
\usepackage{amsmath}
\usepackage{epsfig}
\usepackage{subfigure,multirow}

\title{Learning Sparse Deep Feedforward Networks \\via Tree Skeleton Expansion}

\author{Zhourong Chen, Xiaopeng Li, Nevin L. Zhang\\
Department of Computer Science and Engineering\\
The Hong Kong University of Science and Technology\\
\{zchenbb, xlibo, lzhang\}@cse.ust.hk
}

\begin{document}

\maketitle

\begin{abstract}
Despite the popularity of deep learning, structure learning for deep models remains a relatively under-explored area. In contrast, structure learning has been studied extensively for probabilistic graphical models (PGMs). In particular, an efficient algorithm has been developed for learning a class of tree-structured PGMs called hierarchical latent tree models (HLTMs), where there is a layer of observed variables at the bottom and multiple layers of latent variables on top.  In this paper, we propose a simple method for learning the structures of feedforward neural networks (FNNs) based on HLTMs. 
The idea is to expand the connections in the tree skeletons from HLTMs and to use the resulting structures for FNNs. 
An important characteristic of FNN structures learned this way is that they are sparse. We present extensive empirical results to show that, compared with standard FNNs tuned-manually, sparse FNNs learned by our method achieve better or comparable classification performance with much fewer parameters. They are also more interpretable.
\end{abstract}

\section{Introduction}
Deep learning has achieved great successes in the past few years~\cite{lecun2015deep,hinton2012deepsppech,mikolov2011strategies,krizhevsky2012imagenet}. 
More and more researchers are now starting to investigate the possibility of learning structures for deep models instead of constructing them manually~\cite{chen2017sparse,zoph2016neural,baker2016designing,real2017large}. Structure learning is interesting not only because it can save manual labor, but also because it can yield models that fit data better and hence perform better than manually built ones. In addition, it can also lead to models that are sparse and interpretable.

In this paper, we focus on structure learning for standard feedforward neural networks (FNNs). While convolutional neural networks (CNNs) and recurrent neural networks (RNNs) are designed for spatial and sequential data respectively, standard FNNs are used for data that are neither spatial nor sequential. 
The structures of CNNs and RNNs are relatively more sophisticated than those of FNNs.
For example, a neuron at a convolutional layer in a CNN is connected only to neurons in a small receptive field at the level below. The underlying assumption is that neurons in a small spatial region tend to be strongly correlated in their activations.  In contrast, a neuron in an FNN is connected to all neurons at the level below. We aim to learn sparse FNN structures where a neuron is connected to only a small number of strongly correlated neurons at the level below.

Our work is built upon \emph{hierarchical latent tree analysis} (HLTA)~\cite{DBLP:conf/pkdd/LiuZC14,chen2017latent}, an algorithm for learning tree-structured PGMs where there is a layer of observed variables at the bottom and multiple layers of latent variables on top. HLTA first partitions all the variables into groups such that the variables in each group are strongly correlated and the correlations can be properly modelled using a single latent variable. It then introduces a latent variable for each group. After that it converts the latent variables into observed variables via data completion and repeats the process to produce a hierarchy.

To learn a sparse FNN structure, we assume data are generated from a PGM with multiple layers of latent variables and we try to approximately recover the structure of the generative model. To do so, we first run HLTA to obtain a tree model and use it as a \emph{skeleton}. Then we expand it with additional edges to model salient probabilistic dependencies not captured by the skeleton. The result is a PGM structure and we call it a \emph{PGM core}.   To use the PGM core for classification, we further introduce a small number of neurons for each layer, and we connect them to all the units at the layers and all output units. 
This is to allow features from all layers to contribute to classification directly.

\begin{figure}[t]
\begin{center}
\includegraphics[width=6.5cm]{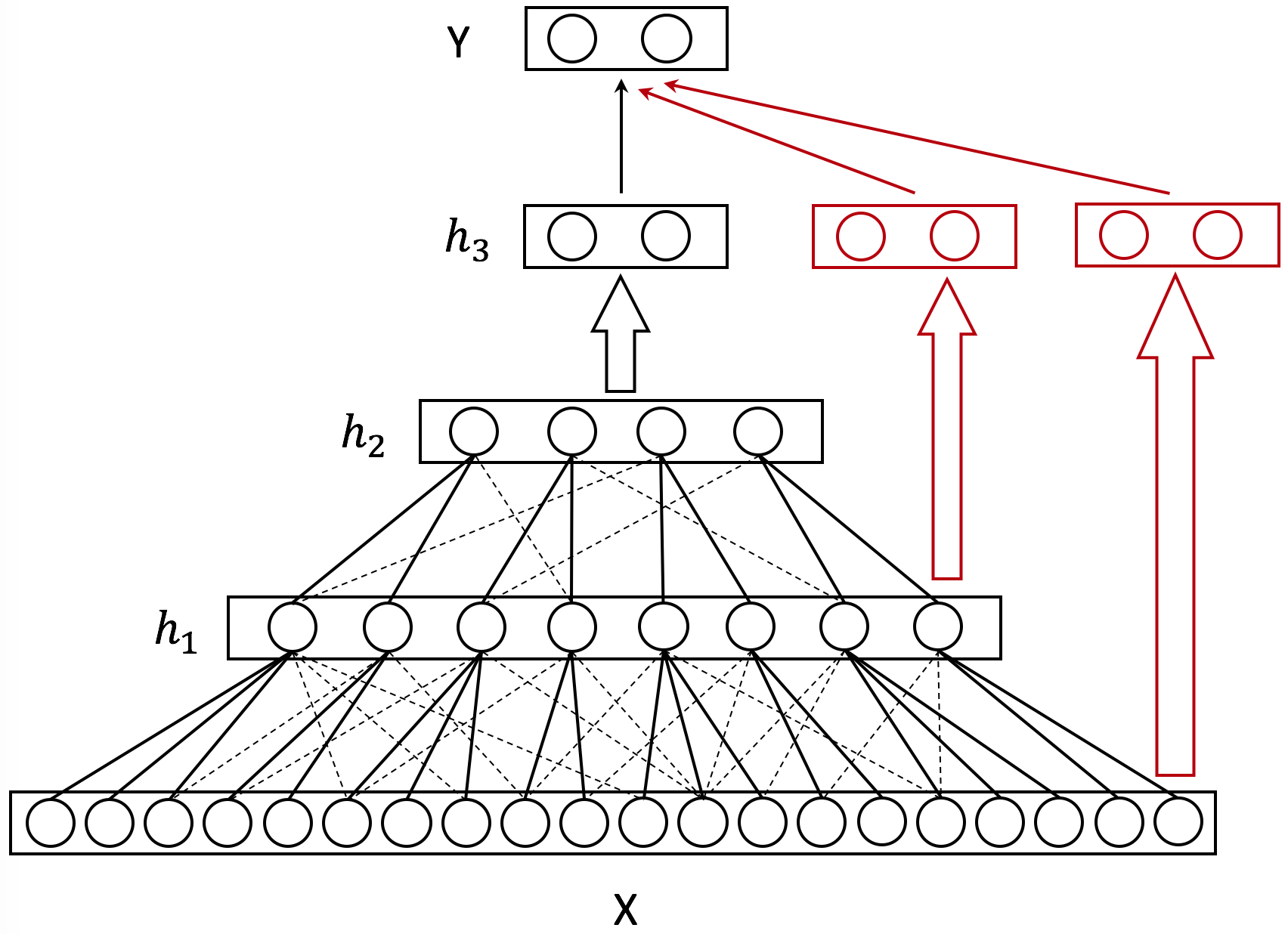}
\end{center}
\vspace{-0.4cm}
\caption{Model structure of our Tree Skeleton Expansion Networks (TSE-Nets). 
The PGM core includes the bottom three layers $x - h_2$. The solid connections make up the skeleton and the dashed connections are added during the expansion phase. 
The black part of the model is called the Backbone, while the red part provides narrow skip-paths from the PGM core to the output layer.}
\vspace{-0.2cm}
\label{fig.BSNN}
\end{figure}

Figure~\ref{fig.BSNN} illustrates the result of our method. The PGM core includes the bottom three layers $x - h_2$. The solid connections make up the skeleton and the dashed connections are added during the expansion phase. The neurons at layer $h_3$ and the output units are added at the last step. The neurons at layer $h_3$ can be conceptually divided into two groups: those connected to the top layer of the PGM core and those connected to other layers. The PGM core, the first group at layer $h_3$ and the output units together form the \emph{Backbone} of the model, while the second group at layer $h_3$ provide narrow \emph{skip-paths} from low layers of the PGM core to the output layer.  
As the structure is obtained by expanding the connections of a tree skeleton, our model is called \emph{Tree Skeleton Expansion Network} (TSE-Net).

Here is a summary of our contributions:
\begin{enumerate}
\item We propose a novel method for learning sparse structures for FNNs. The method depends heavily on HLTA. However, HLTA by itself is not an algorithm for FNN structure learning and it has certainly not been tested for that purpose. 
\item We have conducted extensive experiments to compare TSE-Nets with manually-tuned FNNs.
\item We have analyzed the pros and cons of our method with respect to related works, and we have empirically compared our method with 
a pruning method~\cite{han2015learning} for obtaining FNNs with sparse connectivities.
\end{enumerate}




\section{Related Works}
The primary goal in structure learning is to find a model with optimal 
or close-to-optimal generalization performance. Brute-force search is 
not feasible because the search space is large and evaluating each model 
is costly as it necessitates model training. Early works in the 1980’s and 1990’s have focused on what we call the \emph{micro expansion} approach where 
one starts with a small network and gradually adds new neurons to the 
network until a stopping criterion is met ~\cite{ash1989dynamic,bello1992enhanced,kwok1997constructive}. The word ``micro''
is used here because at each step only one or a few neurons are 
added. This makes learning large model computationally difficult as 
reaching a large model would require many steps and model evaluation is 
needed at each step. In addition, those early methods typically do not 
produce layered structures that are commonly used nowadays. Recently, a 
\emph{macro expansion} method~\cite{liu2017structure} has been proposed where one starts from scratch and repeatedly add layers of hidden units until a threshold is met.

Other recent efforts have concentrated on what we call the \emph{contraction} 
approach where one starts with a larger-than-necessary structure and 
reduces it to the desired size. Contraction can be done either by 
repeatedly pruning neurons and/or connections~\cite{Srinivas2015,li2016pruning,han2015learning}, or by using 
regularization to force some of the weights to zero~\cite{wen2016learning}.  
From the perspective of structure learning, the contraction approach is not 
ideal because it requires a complex model as input. A key motivation for a 
user to consider structure learning is to avoid building models manually.

A third approach is to explore the model space stochastically. One way 
is to place a prior over the space of all possible structures and carry 
out MCMC sampling to obtain a collection of models with high posterior 
probabilities~\cite{adams2010learning}. 
Another way is to encode a model structure as a 
sequence of numbers, use a reinforcement learning meta model to explore the space 
of such sequences, learn a good meta policy from the sequences explored, 
and use the policy to generate model structures~\cite{zoph2016neural}.  
An obvious drawback of such \emph{stochastic exploration} method is that they are 
computationally very expensive.

What we propose in this paper is a \emph{skeleton expansion} method where we 
first learn a tree-structured model and then add a certain number of new 
units and connections to it in one shot. The method has two advantages: 
First, learning tree models is easier than learning non-tree models; 
Second, we need to train only one non-tree model, i.e., the final model.

The skeleton expansion idea has been used in~\cite{chen2017sparse} to learn 
structures for restricted Boltzmann machines, which have only one hidden 
layer. This is the first time that the idea is applied to and tested on 
multi-layer feedforward networks.
\section{Learning Tree Skeleton via HLTA}
\begin{figure*}[t]
\begin{center}
\begin{tabular}{c}
\hspace{-0.8cm}
\mbox{
        \epsfxsize=4.5cm
        \epsffile{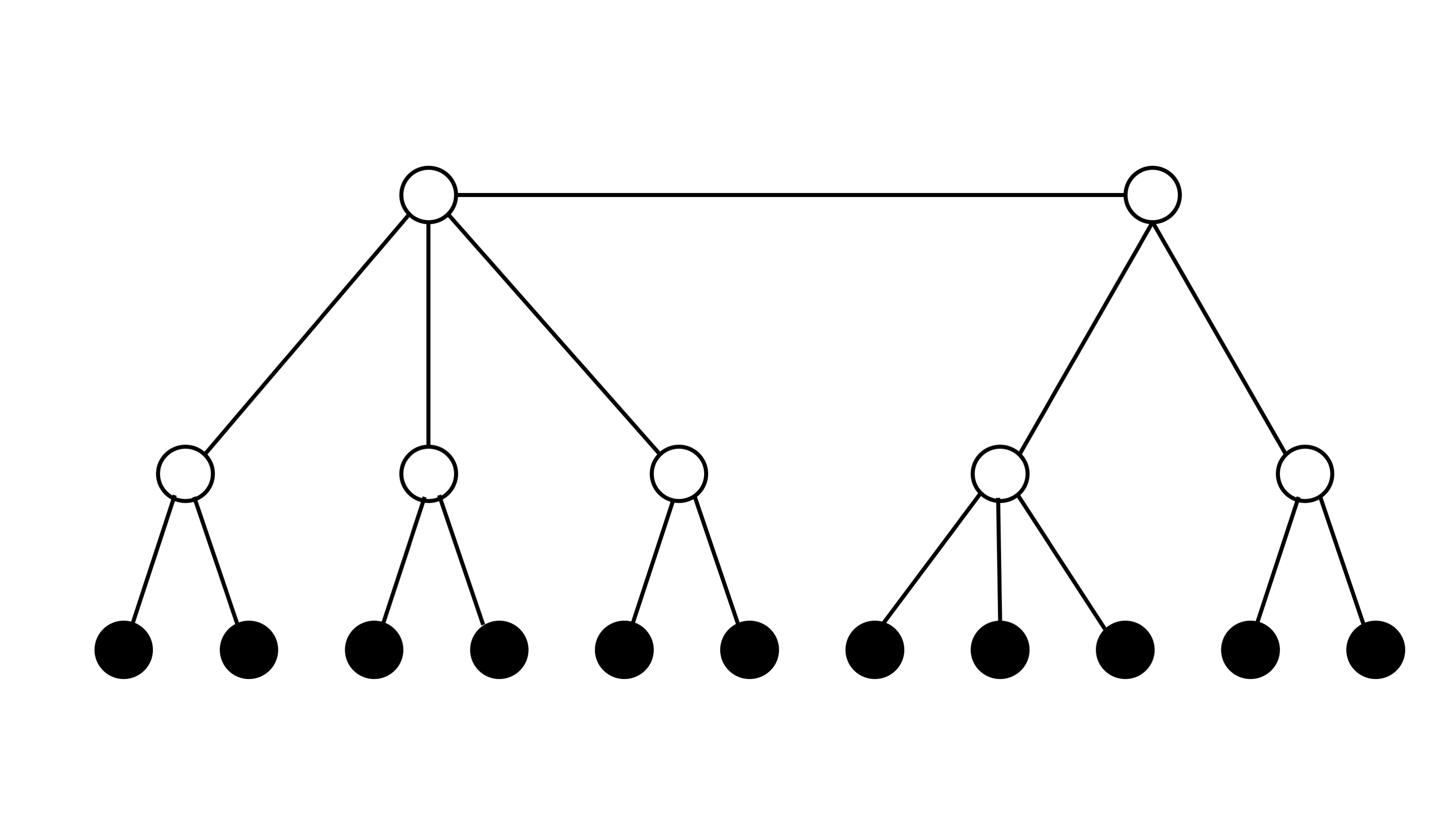}

        \epsfxsize=4.5cm
        \epsffile{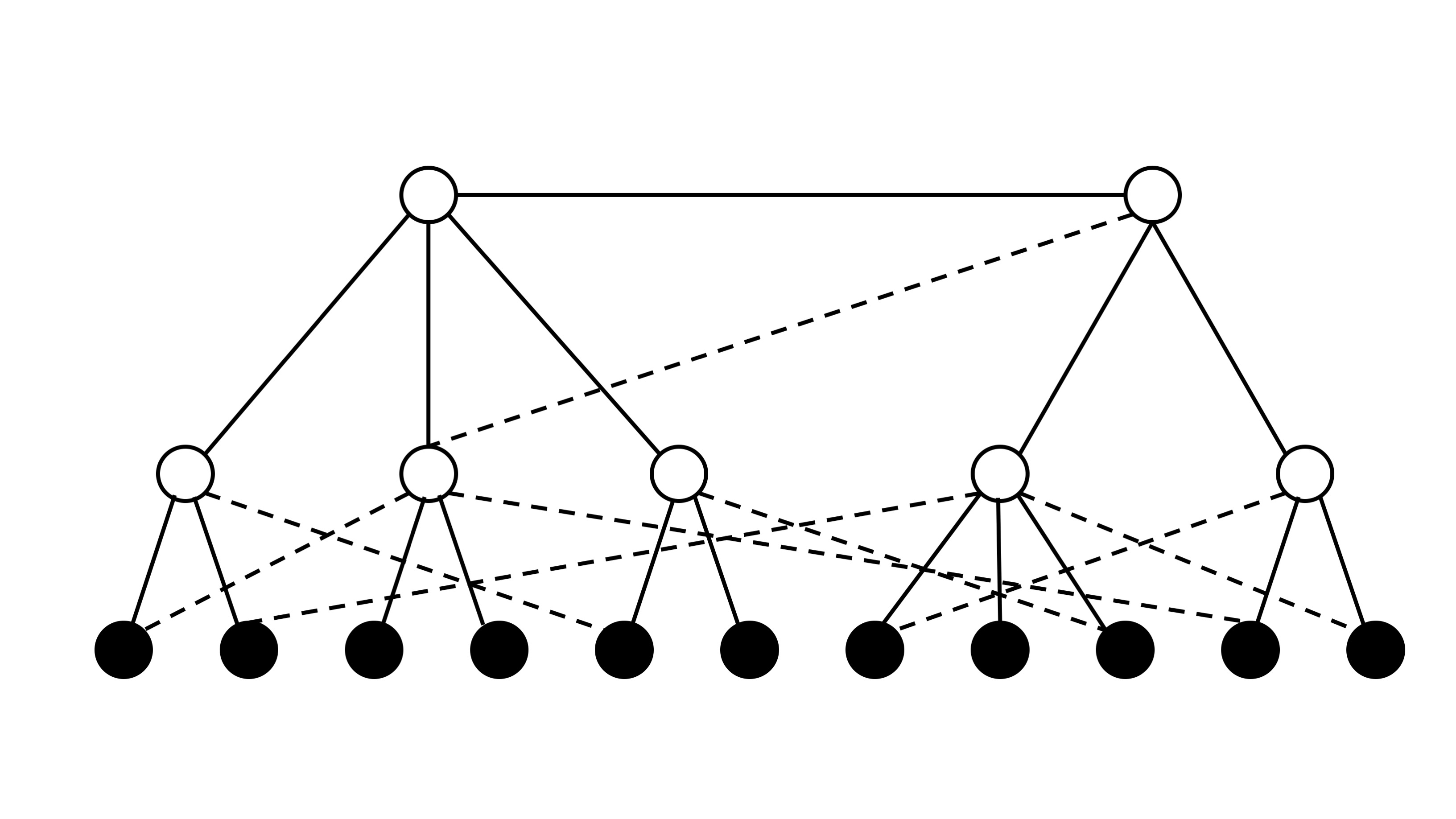}

        \epsfxsize=4.5cm
        \epsffile{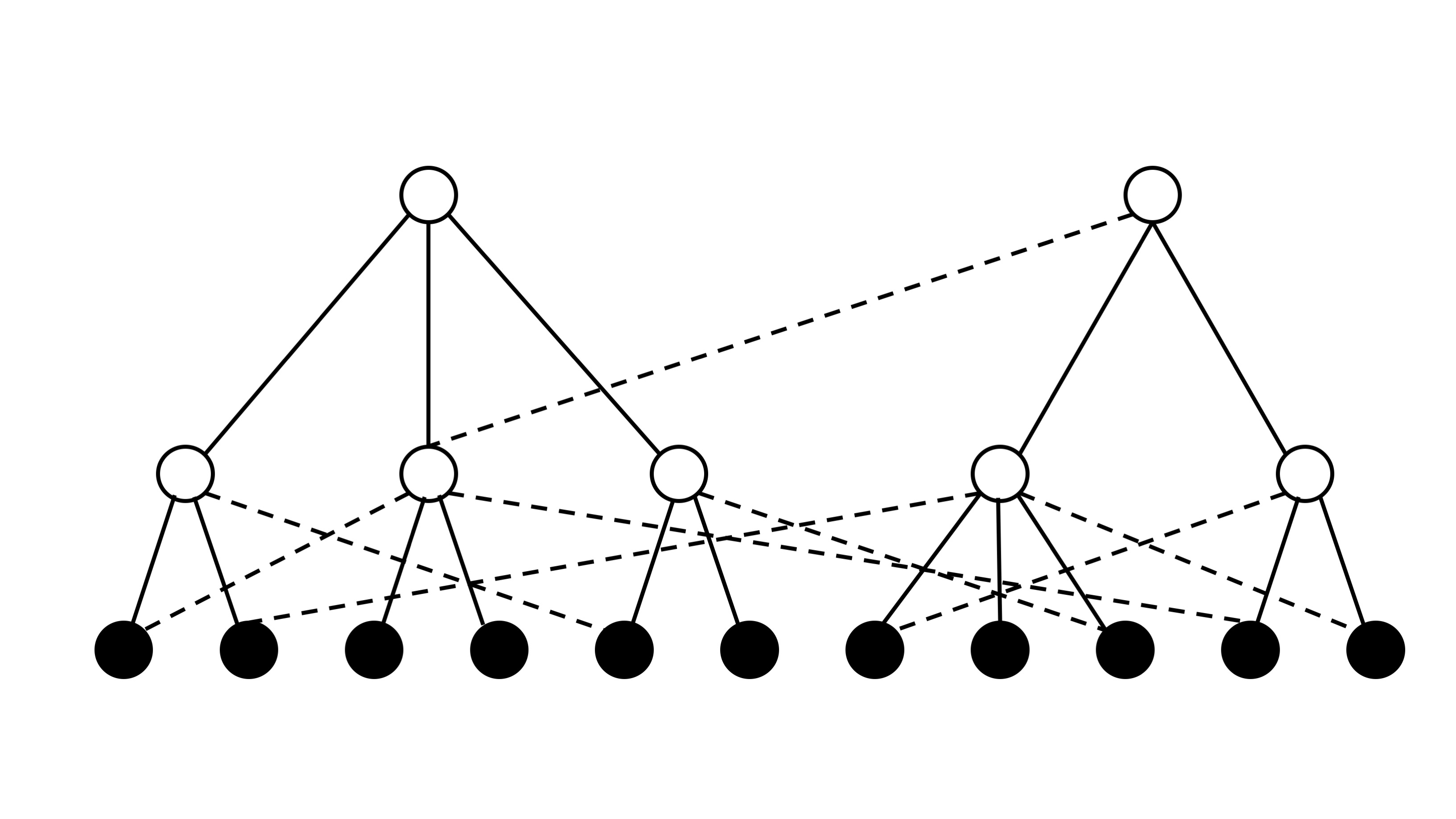}
}
\end{tabular}
\vspace{-0.4cm}
\caption{Tree skeleton expansion: A multi-layer tree skeleton is first learned (left).  New connections are then added to all the layers according to the empirical conditional mutual information (middle). The connections between variables at the top layer are removed and the resulting structure is called the PGM core (right). Black nodes represent observed variables, while white nodes represent latent variables.}
\vspace{-0.3cm}
\label{fig.expansion}
\end{center}
\end{figure*}

The first step of our method is to learn a tree-structured probabilistic graphical model ${\cal T}$ (an example ${\cal T}$ is shown in the left panel in Figure~\ref{fig.expansion}). Let $\mathbf{X}$ be the set of observed variables at the bottom and $\mathbf{H}$ be the unobserved latent variables. 
Then ${\cal T}$ defines a joint distribution over all the variables:
\vspace{-0.15cm}
\begin{small}
\begin{align*}
P(\mathbf{X}, \mathbf{H}) = \prod_{v \in \{\mathbf{X}, \mathbf{H}\}} p(v|pa(v)),
\end{align*}
\end{small}where $pa(v)$ denotes the parent variable of $v$ in ${\cal T}$. The distribution of $\mathbf{X}$ can be computed as:
\vspace{-0.15cm}
\begin{small}
\begin{align*}
P(\mathbf{X}) = \sum_{\mathbf{H}}P(\mathbf{X}, \mathbf{H}).
\end{align*}
\end{small}The model parameters, \begin{small}$\theta = \{p(v|pa(v))\}_{v \in \{\mathbf{X}, \mathbf{H}\}}$\end{small}, can be trained to maximize the data log-likelihood \begin{small}${\log P(D|\theta)}$\end{small} through EM algorithm~\cite{dempster1977maximum}, where \begin{small}$D$\end{small} denotes the data. 
\emph{Stepwise EM}, which is an efficient version of EM similar to stochastic gradient descent, can be used for parameter estimation in ${\cal T}$.

Although parameter learning in ${\cal T}$ is straightforward, the parameterization of ${\cal T}$ depends heavily on the structure of ${\cal T}$ (e.g. how the variables are connected, how many latent variables are introduced) which is relatively difficult to learn. We learn the tree structure of ${\cal T}$ in a layer-wise manner to approximately optimize the BIC score~\cite{schwarz1978estimating} of ${\cal T}$ over data:
\vspace{-0.2cm}
\begin{small}
\begin{align*}
BIC({\cal T}|D) = \log P(D|\theta) - \frac{d}{2}\log(N),
\end{align*}
\end{small}where $d$ denotes the number of free parameters and $N$ is the number of training samples. More specifically, we first learn a two-layer tree structure with $\mathbf{X}$ being the leaf nodes and a layer of latent variables on top. Then we repeat the same process over the layer of latent variables to learn another two-layer tree. In this way, multiple two-layer trees are obtained and we finally stack all the trees up to form a multi-layer tree structure. The whole procedure is shown in Figure~\ref{fig.process}.

\subsection{Learning A Two-Layer Tree}
\begin{figure}
\begin{center}
\includegraphics[width=6cm]{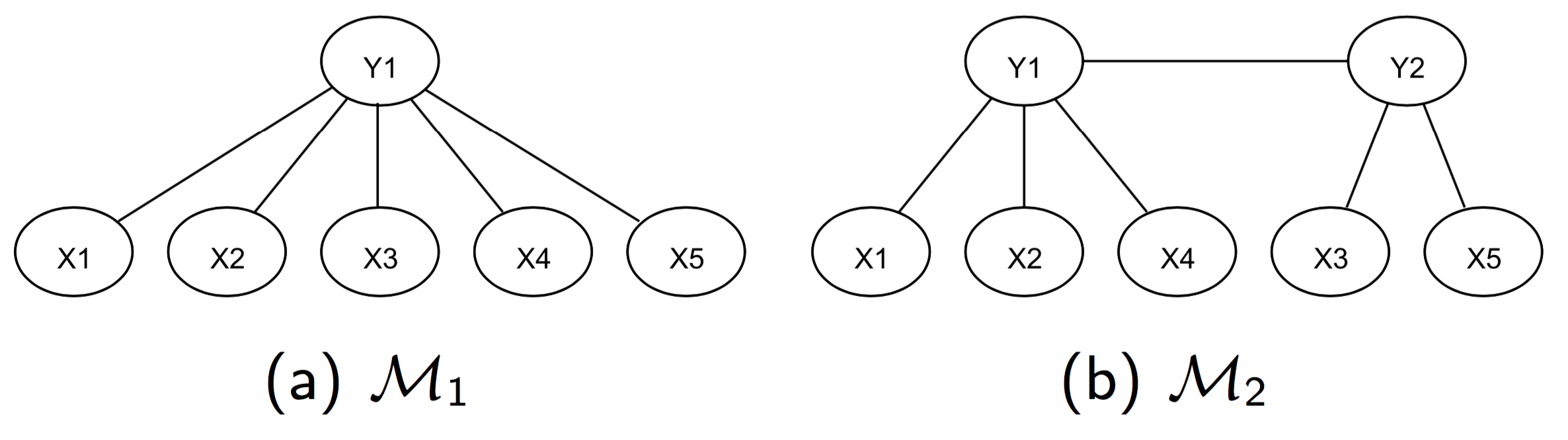}
\end{center}
\vspace{-0.5cm}
\caption{Example of testing whether five observed variables should be grouped together: (a) The best model with one latent variable. (b) The best model with two latent variables.}
\vspace{-0.3cm}
\label{fig.M1M2}
\end{figure}

To learn a two-layer tree structure, we need to first partition the observed variables into disjoint groups such that each group are strongly correlated and can be placed together as children of a shared latent variable. 
HLTA achieves this by greedily optimizing the BIC score of the model.
It starts by finding the two most correlated variables to form one group and keeps expanding the group if necessary. Let $S$ denotes the set of observed variables which haven't been included into any variable groups.
HLTA firstly computes the mutual information between each pair of observed variables. Then it picks the pair in $S$ with the highest mutual information and uses them as the seeds of a new variable group $G$. 
New variables from $S$ are then added to $G$ one by one in descending order of their mutual information with variables already in $G$. 
Each time when a new variable is added into $G$, HLTA builds two models (${\cal M}_1$ and ${\cal M}_2$) with $G$ as the observed variables. The two models are the best models with one single latent variable and two latent variables respectively, as shown in Figure~\ref{fig.M1M2}. HLTA computes the BIC scores of the two models and tests whether the following condition is met:
\begin{small}
$$BIC({\cal M}_2|D) - BIC({\cal M}_1|D) \le \delta, $$
\end{small}where $\delta$ is a threshold which is usually set at 3~\cite{chen2017latent}. 
When the condition is met, the two latent variable model ${\cal M}_2$ is not significantly better than the one latent variable model ${\cal M}_1$. Correlations among variables in $G$ are still well modelled using a single latent variable. Then HLTA keeps on adding new variables to $G$. If the test fails, HLTA takes the subtree in ${\cal M}_2$ which doesn't contain the newly added variable and identifies the observed variables in it as a finalized variable group. The group are then removed from $S$. And the above process is repeated on $S$ until all the variables in $S$ are partitioned into disjoint groups. An efficient algorithm, \emph{Progressive EM}~\cite{chen2016progressive}, is used to estimate the parameters in ${\cal M}_1$ and ${\cal M}_2$.

After partitioning the observed variables into groups, we introduce a latent variable for each group and compute the mutual information among the latent variables. Then we link up the latent variables to form a Chow-Liu Tree~\cite{chow1968approximating} based on their mutual information. The result is a latent tree model ~\cite{Pearl:1988:PRI:52121,zhang2004hierarchical}, as shown in Figure~\ref{fig.process}(c). Note that all the latent variables here are directly connected to some observed variables and hence we call them the layer-1 latent variables.

\subsection{Stacking Two-Layer Trees to a Tree Skeleton}
After learning the first two-layer tree, we convert the layer-1 latent variables to observed variables $\mathbf{X}^{\prime}$ through data completion. 
Then another two-layer tree can be built by applying the above method to $\mathbf{X}^{\prime}$. The resulting tree gives the layer-2 latent variables. 
And this procedure of building a two-layer tree can be repeated on the newly-introduced latent variables recursively until the number of latent variables on top falls below a threshold $T$, resulting in multiple two-layer trees.
Finally we stack all the two-layer trees up one by one to form a tree skeleton ${\cal T}$ with multiple layers of latent variables as shown in Figure~\ref{fig.process}(e). Note that the connections between latent variables at the same layer are removed, except those in the top layer. The result here is a hierarchical latent tree model~\cite{DBLP:conf/pkdd/LiuZC14,chen2017latent}.

{
\begin{figure}
\begin{center}
\includegraphics[width=4cm]{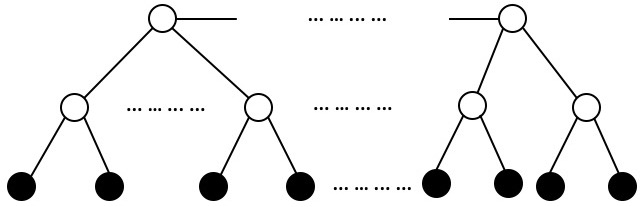} \\  (e) \\
\vspace{0.2cm}

\includegraphics[width=4cm]{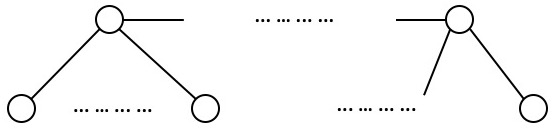} \\  (d) \\
\vspace{0.2cm}

\includegraphics[width=4cm]{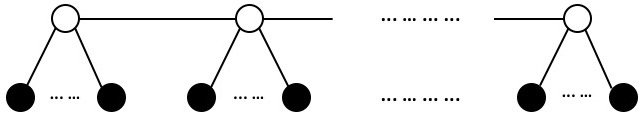} \\  (c) \\ 
\vspace{0.2cm}

\includegraphics[width=4cm]{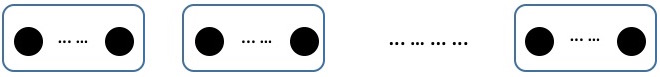} \\  (b) \\
\vspace{0.2cm}

\includegraphics[width=4cm]{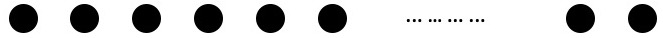} \\  (a)

\end{center}
\vspace{-0.4cm}
\caption{The structure learning procedure for multi-layer tree skeleton. Black nodes represent observed variables while white nodes represent latent variables. (a) A set of observed variables. (b) Partition the observed variables into groups. (c) Introduce a latent variable for each group and link the latent variables up as a Chow-Liu tree. (d) Convert the layer-1 latent variables into observed variables and repeat the previous process on them to obtain another two-layer tree. (e) Stack the multiple two-layer trees up to form a multi-layer tree skeleton.}
\vspace{-0.2cm}
\label{fig.process}
\end{figure}
}

\section{Expanding Tree Skeleton to PGM Core}
We have restricted the structure of ${\cal T}$ to be a tree, as parameter estimation in tree-structured PGMs is relatively efficient. However, this restriction in return also hurts the model's expressiveness. 
For example, in text analysis, the word {\it Apple} is highly correlated with both fruit words and technology words conceptually. But {\it Apple} is directly connected to only one latent variable in ${\cal T}$ and it is difficult for the latent variable to express both the two concepts, which may cause severe underfitting. 
On the other hand, in standard FNNs, units at a layer are always fully connected to those at the previous layer, resulting in high connection redundancies. 

In this paper, we aim to learn sparse connections between adjacent layers, such that they are neither as sparse as those in a tree, nor as dense as those in an FNN. 
To this end, the sparse connections should capture only the most important correlations among the observed variables. Thus we propose to use ${\cal T}$ as a structure skeleton and expand it to a denser structure ${\cal G}$ which we call the \emph{PGM core}. 

Let $V_l$ be a latent variable at layer $l$ in ${\cal T}$. We consider adding new connections to link it to more variables at layer $l-1$. We evaluate the importance of a connection by computing the empirical conditional mutual information:
\begin{scriptsize}
\begin{align*}
\hat I(V_{l}, V_{l-1} | Z) = \sum_{Z}\hat p(Z)\sum_{V_l}\sum_{V_{l-1}}\hat p(V_{l},V_{l-1}|Z)\log \frac{\hat p(V_{l}, V_{l-1}|Z)}{\hat p(V_{l}|Z) \hat p(V_{l-1}|Z)},
\end{align*}
\end{scriptsize}where $Z$ is the parent of $V_{l-1}$ in ${\cal T}$ and $Z\neq V_l$. If ${\cal T}$ with its tree structure perfectly models the correlation between $V_{l}$ and $V_{l-1}$, then $V_{l}$ and $V_{l-1}$ should be conditional independent and thus $\hat I(V_{l}, V_{l-1} | Z)$ should be zero. In other words, if $\hat I(V_{l}, V_{l-1} | Z)$ is a large value other than zero, then it indicates that the correlation between $V_{l}$ and $V_{l-1}$ is still not well modelled by ${\cal T}$ and we should consider adding a new connection between $V_{l}$ and $V_{l-1}$. With this intuition, we sort $V_{l-1}$ in descending order by $\hat I(V_{l}, V_{l-1} | Z)$, and add new links to connect the top $K$ variables $V_{l-1}$ to $V_{l}$. This expansion phase is carried out over all the adjacent layers. 
We then remove the connections between the top layer latent variables and call the resulting structure the \emph{PGM core} ${\cal G}$. The skeleton expansion phase is shown in Figure~\ref{fig.expansion}. Note that at this stage we don't need to learn the parameters of the new connections.

\section{Constructing Sparse FNNs from PGM Core}
Our tree expansion method learns a multi-layer sparse structure ${\cal G}$ in an unsupervised manner. 
One key advantage of unsupervised structure learning is, the structure learned from a set of unlabelled data can be transfered to any supervised learning tasks on the same type of data. 
Convolutional layer widely used in computer vision tasks is a good example:
We humans have seen many unlabelled scenes and conclude that there are strong correlations between neighbouring pixels in vision data. And hence we humans design the locally-connected structure of convolutional layer which is well suited to the nature of vision data and works well in supervised learning tasks.
Similarly, our method discovers strong correlations in general data other than images. To utilize the resulting structure in a discriminative model, we convert each latent variable $h$ in ${\cal G}$ to a hidden unit by defining the conditional probability:
\begin{small}
\begin{align*}
p(h|\mathbf{x}) = o(\mathbf{W'x}+b),
\end{align*}
\end{small}where $\mathbf{x}$ denotes a vector of the units directly connected to $h$ at the layer below, $\mathbf{W}$ and $b$ are connection weights and bias respectively, and $o$ denotes a probability function mapping real values to probabilities, e.g. the sigmoid function. As in many deep learning models, we can further replace $o$ with other non-linear activation functions, such as the tanh and ReLU~\cite{nair2010rectified,glorot2011deep} functions which usually benefit the training of deep models. In this way, we convert ${\cal G}$ into a sparse multi-layer neural network. Next we discuss how we use it as a feature extractor in supervised learning tasks. Our model contains two parts, the \emph{Backbone} and the \emph{skip-paths}.\vspace{-0.5cm}
\paragraph{The Backbone}For a specific classification or regression task, we introduce a fully-connected layer on the top of ${\cal G}$, which we call the \emph{feature layer}, followed by a output layer. 
As shown in Figure~\ref{fig.BSNN}, the feature layer acts as a feature ``aggregator'', aggregating the features extracted by ${\cal G}$ and feeding them to the output layer. 
We call the whole resulting module (${\cal G}$, feature layer and output layer together) the Backbone, as it is supposed to be the major module of our model.\vspace{-0.5cm}
\paragraph{The Skip-paths}As the structure of ${\cal G}$ is sparse and is learned to capture the strongest correlations in data, some weak but useful correlations may easily be missed. More importantly, different tasks may rely on different weak correlations and this cannot be taken into consideration during the unsupervised structure learning.
To remedy this, we consider allowing the model to contain some narrow fully-connected paths to the feature layer such that they can capture those missed features.
More specifically, suppose there are $L$ layers of units in ${\cal G}$. We introduce $L-1$ more groups of units into the feature layer, with each group fully connected to a layer in ${\cal G}$ (except the top layer). In this way, each layer except the top one in ${\cal G}$ has both a sparse path (the Backbone) and a fully-connected path to the feature layer. The fully-connected paths are supposed to capture those minor features during parameter learning. These new paths are called \emph{skip-paths}.

As shown in Figure~\ref{fig.BSNN}, the Backbone and the skip-paths together form our final model, named \emph{Tree Skeleton Expansion Network} (TSE-Net). The model can then be trained like a normal neural network using back-propagation.

\section{Experiments}
\subsection{Datasets}
We evaluate our method in 17 classification tasks. Table~\ref{table.dataset} gives a summary of the datasets. We choose 12 tasks of chemical compounds classification and 5 tasks of text classification. All the datasets are published by previous researchers and are available to the public.\vspace{-0.5cm}
\paragraph{Tox21 challenge dataset}\footnote{https://github.com/bioinf-jku/SNNs} 
There are about 12,000 environmental chemical compounds in the dataset, each represented as its chemical structure. The tasks are to predict 12 different toxic effects for the chemical compounds. We treat them as 12 binary classification tasks. We filter out sparse features which are present in fewer than 5\% compounds, and rescale the remaining 1,644 features to zero mean and unit variance. The validation set is randomly sampled and removed from the original training set.\vspace{-0.5cm}
\paragraph{Text classification datasets}\footnote{https://github.com/zhangxiangxiao/Crepe}
We use 5 text classification datasets from~\cite{zhang2015character}. After removing stop words, the top 10,000 frequent words in each dataset are selected as the vocabulary respectively and each document is represented as bag-of-words over the vocabulary. The validation set is randomly sampled and removed from the training samples.
\begin{table}
\caption{Statistics of all the datasets.}
\vspace{-0.4cm}
\label{table.dataset}
\begin{center}
\begin{small}
\begin{tabular}{llrrr}
Dataset           & Classes & \begin{tabular}[c]{@{}l@{}}Training \\ Samples\end{tabular} & \begin{tabular}[c]{@{}l@{}}Validation \\ Samples\end{tabular} & \begin{tabular}[c]{@{}l@{}}Test \\ Samples\end{tabular} \\ \hline
Tox21             & 2       & $\sim$~9,000  			 & 500       & $\sim$600  \\
Yelp Review Full  & 5       & 640,000            & 10,000             & 50,000         \\
DBPedia           & 14      & 549,990            & 10,010             & 70,000         \\
Sogou News        & 5       & 440,000            & 10,000             & 60,000         \\
Yahoo!Answer      & 10      & 1,390,000          & 10,000             & 60,000         \\
AG's News         & 4       & 110,000            & 10,000             & 7,600         
\end{tabular}
\vspace{-0.2cm}
\end{small}
\end{center}
\end{table}

\begin{table}[t]
\caption{Hyper-parameters for the structure of FNNs.}
\vspace{-0.4cm}
\label{table.FNN-hyper}
\begin{center}
\begin{small}
\begin{tabular}{ll}
Hyper-parameter         & Values considered    \\ \hline
Number of units per layer      & \{512, 1024, 2048\}  \\
Number of hidden layers & \{1,2,3,4\}          \\
Network shape           & \{Rectangle, Conic\}
\end{tabular}
\vspace{-0.4cm}
\end{small}
\end{center}
\end{table}


\subsection{Experiment Setup}
We compare our model TSE-Net with standard FNN.
For fair comparison, we treat the number of units and number of layers as hyper-parameters of an FNN and optimize them via grid-search over all the defined combinations using validation data. Table~\ref{table.FNN-hyper} shows the space of network configurations considered, following the setup in~\cite{klambauer2017self}. In our TSE-Net, the number of layers and the number of units at each layer are determined by the algorithm. We set the upper-bound $T$ for the number of units at the top layer in ${\cal G}$ to around 500, resulting in ${\cal G}$ with 2 or 3 hidden layers. In the expansion phase, we expand the connections such that each unit in ${\cal G}$ is connected to 5\% of the units at the layer below. 
By sampling a subset of data for structure learning, our method runs efficiently on a standard desktop.

We also compare our model with pruned FNN whose connections are sparse.
We take the best FNN as the initial model and perform pruning as in~\cite{han2015learning}. As micro expansion and stochastic exploration methods are not learning layered FNNs and are computationally expensive, they are not included in comparison.

We use ReLUs~\cite{nair2010rectified} as the non-linear activation functions in all the networks. Dropout~\cite{hinton2012improving,srivastava2014dropout} with rate 0.5 is applied after each non-linear projection. We use Adam~\cite{kingma2014adam} as the network optimizer. 
Codes will be released after the paper is accepted.

\subsection{Results}
\begin{figure}[t]
\begin{center}
\begin{tabular}{c}
\hspace{-0.4cm}
\mbox{
        \epsfxsize=4cm
        \epsffile{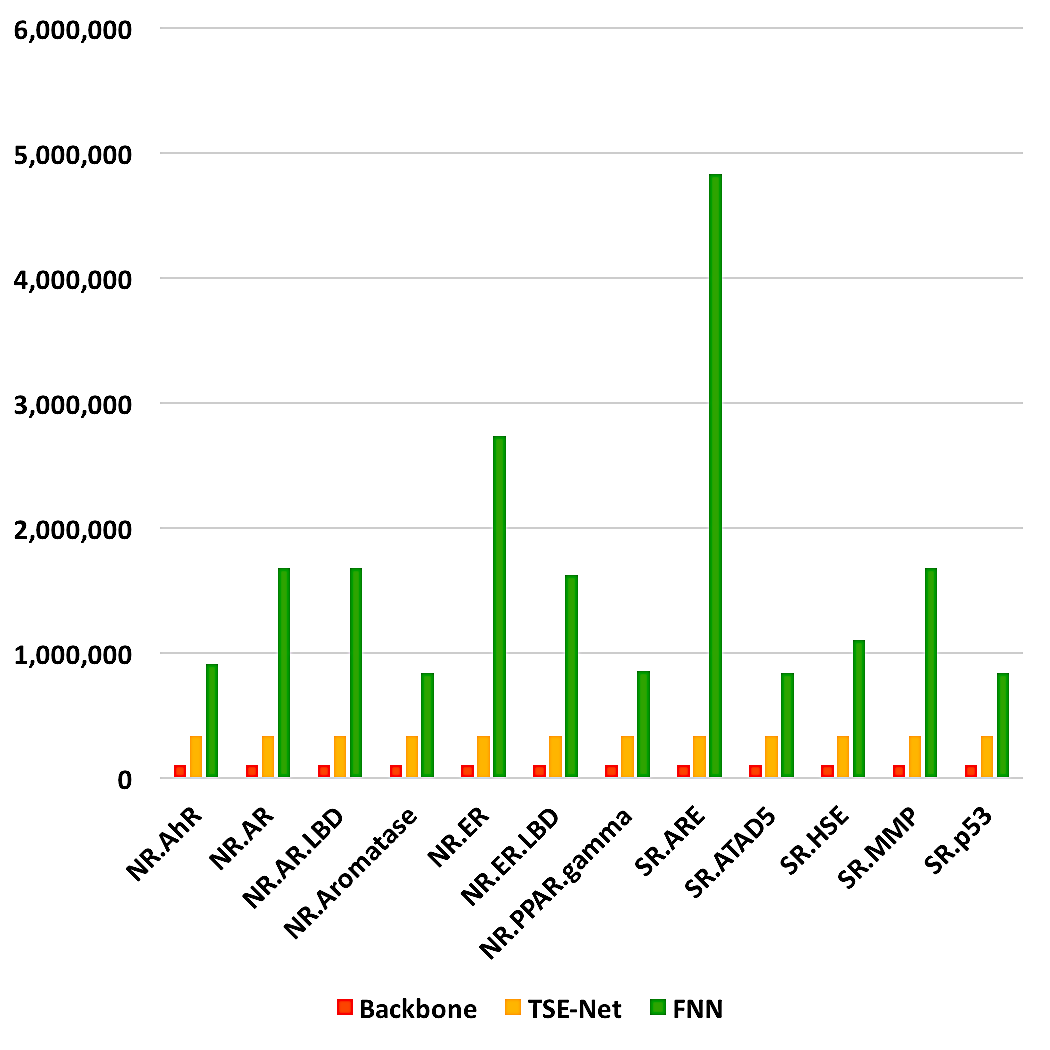}

        \hspace{0.1cm}
        \epsfxsize=4.5cm
        \epsffile{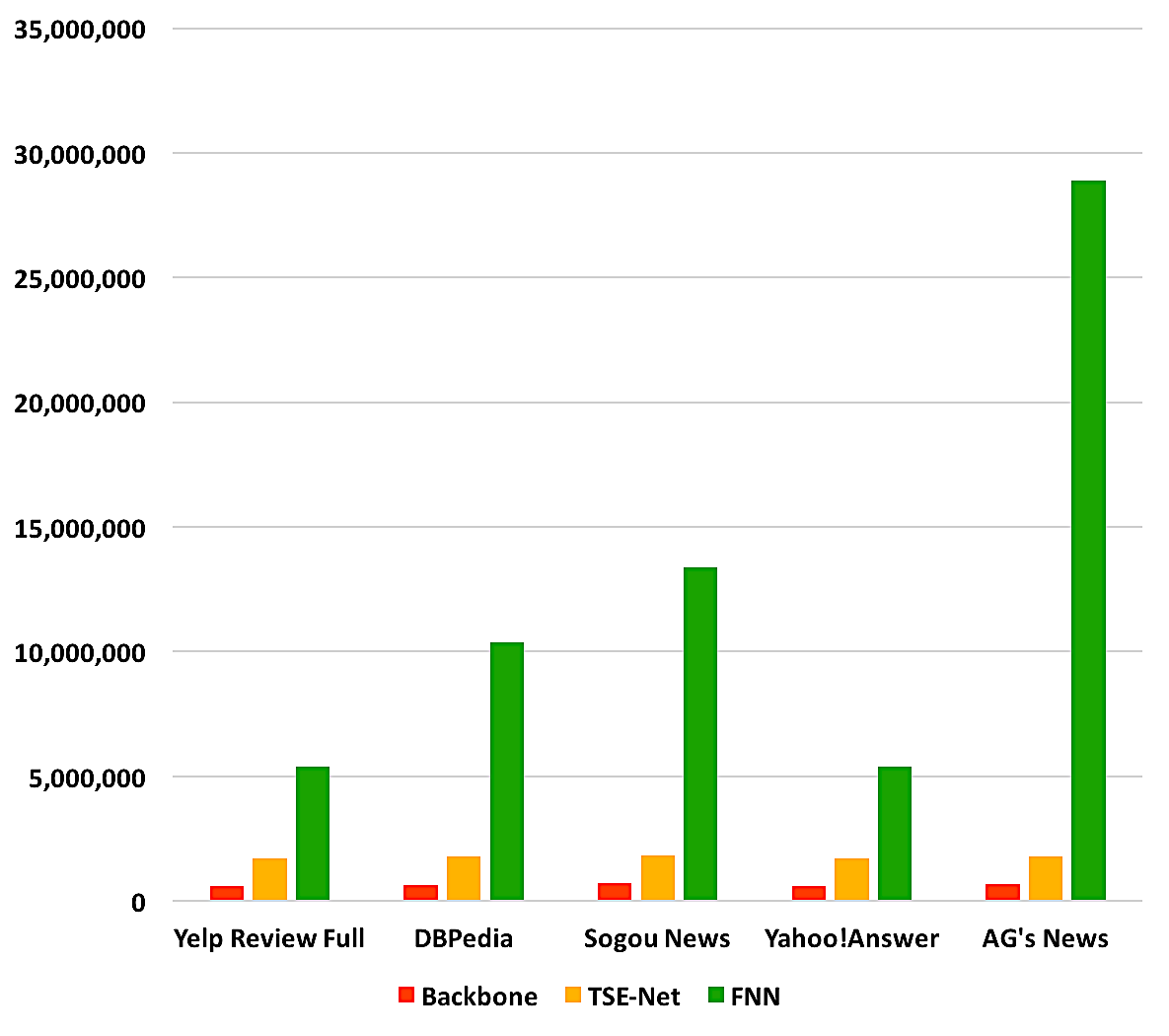}
}
\end{tabular}
\caption{The number of parameters in TSE-Nets and FNNs for different tasks. The left panel shows the 12 tasks in the Tox21 dataset.}
\label{fig.parameter}
\vspace{-0.4cm}
\end{center}
\end{figure}

Classification results are reported in Table~\ref{table.result}. All the experiments are run for three times and we report the average classification AUC scores/accuracies with standard deviations.\vspace{-0.5cm}
\paragraph{TSE-Nets vs FNNs}
From the table we can see that, TSE-Net contains only 6.25\%$\sim$32.07\% of the parameters in FNN.
Although the structure of FNN is manually optimized over the validation data, TSE-Net still achieves better or comparable results than FNN with much fewer parameters.
In our experiments, TSE-Net achieves better AUC scores than FNN in 10 out of the 12 tasks in Tox21 dataset.
The results show that, although TSE-Net is much sparser than FNN, the structure successfully captures the crucial correlations in data and greatly reduces the number of parameters without significant performance loss. The number of parameters in different models are also plotted in Figure~\ref{fig.parameter}, which clearly shows that TSE-Nets contain much fewer parameters than FNNs.

It is worth noting that pure FNNs are not the state-of-the-art models for the tasks here. For example, \cite{mayr2016deeptox} proposes an ensemble of FNNs, random forests and SVMs with expert knowledge for the Tox21 dataset. \cite{klambauer2017self} tests different normalization techniques for FNNs on the Tox21 dataset. They both achieve an average AUC score around 0.846. Complicated RNNs~\cite{yang2016hierarchical} with attention also achieve better results than FNNs for the 5 text datasets. However, the goal of our paper is to improve standard FNNs by learning sparse structure, instead of proposing state-of-the-art methods for any specific tasks. Their methods are all much more complex and even task-specific, and hence it is not fair to include their results as comparison. Moreover, their methods can also be combined with our TSE-Nets to give better results.\vspace{-0.5cm}
\paragraph{Contribution of the Backbone}
To validate our assumption that the Backbone in TSE-Net captures most of the crucial correlations in data and acts as a main part of the model, we remove the narrow skip-paths in TSE-Net and train the model to test its performance. 
\begin{table*}[t]
\centering
\caption{Experiment results. For each model, the first row shows the classification AUC scores/accuracies, while the second row shows the number of parameters and the ratio w.r.t that of FNNs. Best result on each dataset is marked in \textbf{bold}.}
\label{table.result}
\begin{small}
\begin{tabular}{l|llllll}
                              & Tox21 Average                & Yelp Review                   & DBPedia                       & Sogou News                    & Yahoo!Answer                  & AG's News                     \\ \hline
\multirow{2}{*}{FNN}          & 0.8010 $\pm$ 0.0017          & 59.13\% $\pm$ 0.14\%          & 97.99\% $\pm$ 0.04\%          & 96.12\% $\pm$ 0.06\%          & \textbf{71.84\% $\pm$ 0.07\%} & \textbf{91.61\% $\pm$ 0.01\%} \\
                              & 1.64M                        & 5.38M                         & 10.36M                        & 13.39M                        & 5.39M                         & 28.88M                        \\ \hline
\multirow{2}{*}{TSE-Net}      & \textbf{0.8150 $\pm$ 0.0038} & \textbf{59.14\% $\pm$ 0.06\%} & \textbf{98.11\% $\pm$ 0.03\%} & 96.09\% $\pm$ 0.06\%          & 71.42\% $\pm$ 0.06\%          & 91.39\% $\pm$ 0.03\%          \\
                              & 338K/20.64\%                 & 1.73M/32.07\%                 & 1.78M/17.13\%                 & 1.84M/13.77\%                 & 1.69M/31.42\%                 & 1.81M/6.25\%                  \\ \hline
\multirow{2}{*}{Backbone} & 0.7839 $\pm$ 0.0076          & 58.63\% $\pm$ 0.13\%          & 97.91\% $\pm$ 0.04\%          & 95.67\% $\pm$ 0.04\%          & 69.95\% $\pm$ 0.08\%          & 91.33\% $\pm$ 0.03\%          \\
                              & 103K/6.29\%                  & 613K/11.38\%                  & 651K/6.28\%                   & 712K/5.32\%                   & 582K/10.80\%                  & 678K/2.35\%                   \\ \hline
Pruned FNN                    & 0.7998 $\pm$ 0.0034          & 59.12\% $\pm$ 0.01\%          & \textbf{98.11\% $\pm$ 0.02\%} & \textbf{96.20\% $\pm$ 0.06\%} & 71.74\% $\pm$ 0.05\%          & 91.49\% $\pm$ 0.09\%         
\end{tabular}
\end{small}
\end{table*}
As we can see from the results, the Backbone path alone already achieves AUC scores or accuracies which are only slightly worse than those of TSE-Net. 
Note that the number of parameters in the Backbone is even much smaller than that of TSE-Net. The Backbone contains only 2\%$\sim$11\% of the parameters in FNN. 
The results not only show the importance of the Backbone in TSE-Net, but also show that our structure learning method for the Backbone path is effective.\vspace{-0.5cm}
\paragraph{TSE-Nets vs Pruned FNNs}
We also compare our method with a baseline method~\cite{han2015learning} for obtaining sparse FNNs. The pruning method provides regularization over the weights of a network. The regularization is even stronger than $l1/l2$ norm as it is producing many weights being exactly zeros. We start from the fully pretrained FNNs reported in Table~\ref{table.result}, and prune the weak connections with the smallest absolute weight values. The pruned networks are then retrained again to compensate for the removed connections. After pruning, the number of remaining parameters in each FNN is the same as that in the corresponding TSE-Net for the same task. As shown in Table~\ref{table.result}, TSE-Net and pruned FNN achieve pretty similar results. Without any supervision or pre-training over connection weights, our unsupervised structure learning method successfully identifies important connections and learns sparse structures. This again validates that our method is effective for learning sparse structures.\vspace{-0.5cm}
\paragraph{Interpretability}
We also compare the interpretability of different models on the text datasets following the experiments in~\cite{chen2017sparse}. We feed the data to the networks and conduct forward propagation to get the values of the hidden units corresponding to each data sample. Then for each hidden unit, we sort the words in descending order of the correlations between the words and the hidden unit. The top 10 words with the highest correlations are chosen to characterize the hidden unit. 
Following \cite{chen2017sparse}, we measure the “interpretability” of a hidden unit by considering how similar each pair of words in the top-10 list are. 
The similarity between two words is calculated from a word2vec model \cite{mikolov2013efficient,DBLP:conf/nips/MikolovSCCD13} trained on the Google News datasets and released by Google, where each word is mapped to a high dimensional vector. The similarity between two words is defined as the cosine similarity of the two corresponding vectors. The interpretability score of a hidden unit is computed as the average similarity of all pairs of words. And the interpretability score of a model is defined as the average interpretability score of all hidden units.

Table~\ref{table.interpretscore} reports the interpretability scores of TSE-Nets, FNNs and Pruned FNNs. \textit{Sogounews} dataset is not included in the experiment since its vocabulary are Chinese pingyin characters and most of them do not appear in the Google News word2vec model. We measure the interpretability scores by considering only the top-layer hidden units. 
From the table we can see that, TSE-Nets significantly outperform FNNs and Pruned FNNs in most cases and is comparable if not better, showing superior coherency and compactness in the characterizations of hidden units and thus better model interpretability.

\begin{table}[t]
\caption{Interpretability scores of TSE-Nets, FNNs and Pruned FNNs on different datasets}
\vspace{-0.2cm}
\label{table.interpretscore}
\begin{center}
\begin{small}
\begin{tabular}{llll}
Task & TSE-Nets & FNNs & Pruned FNNs \\
\hline
Yelp Review Full 	&  \textbf{0.1632} &  0.1117 & 0.1000\\
DBPedia				&  \textbf{0.0609} &  0.0497 & 0.0553\\
Yahoo!Answer		&  \textbf{0.1729} &  0.1632 & 0.1553\\
AG's News 			&  0.0531 & \textbf{0.0595} & 0.0561\\
\end{tabular}
\end{small}
\end{center}
\vspace{-0.4cm}
\end{table}

\begin{figure}[h]
\begin{center}
\begin{tabular}{c}
\hspace{-0.4cm}
\mbox{
        \epsfxsize=2cm
        \epsffile{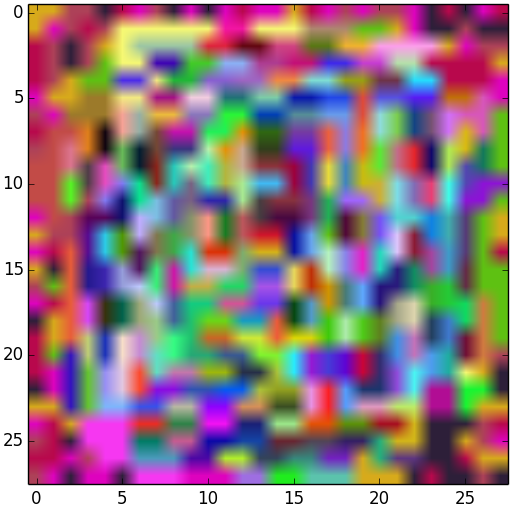}

        \hspace{0.1cm}
        \epsfxsize=2cm
        \epsffile{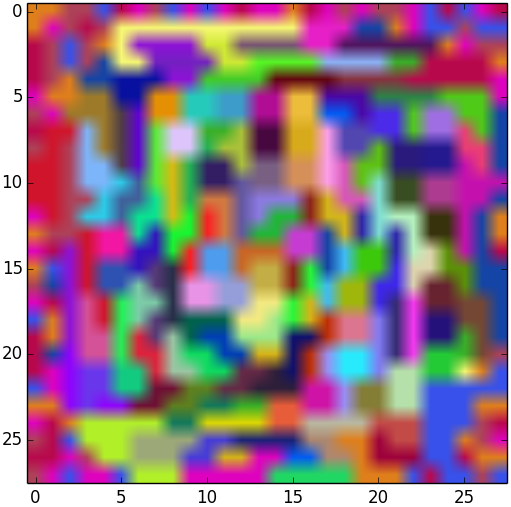}

        \hspace{0.1cm}
        \epsfxsize=2cm
        \epsffile{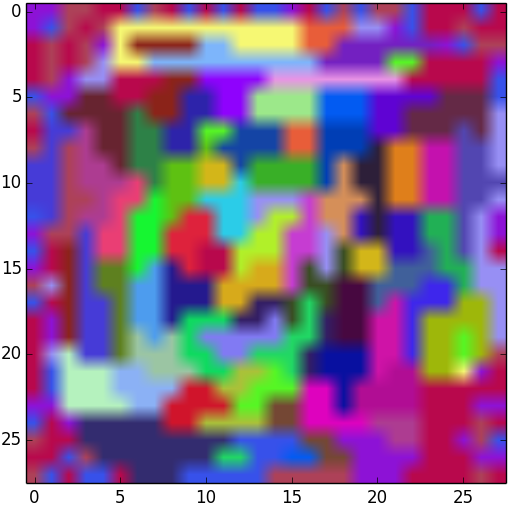}
}
\end{tabular}
\caption{The partitions of the MNIST pixels by layer-1, layer-2 and layer-3 latent variables respectively. Pixels in each sub-figure with the same color belong to the same group.}
\label{fig.mnist}
\vspace{-0.6cm}
\end{center}
\end{figure}

To further demonstrate that our method can learn good structures, we apply it to the MNIST dataset~\cite{lecun1998gradient} to learn a tree skeleton. Each layer of latent variables partition the pixels into disjoint groups. We take the first three layers of latent variables and visualize the partitions of pixels in Figure~\ref{fig.mnist}. In each sub-figure, pixels with the same color belong to the same group. 
As we can see, even though pixel location information is not used in the analysis, our method grouped neighboring pixels together. The reason is that neighbor pixels tend to be strongly correlated. Also note that the blocks at the bottom and the top are mostly horizontal, while those in the middle are often vertical. Those reflect the interesting characteristics of handwritten digits. 

\section{Conclusions}
Structure learning for deep neural network is a challenging and interesting research problem. We have proposed an unsupervised structure learning method which utilizes the correlation information in data for learning sparse deep feed-forward networks. 
In comparison with standard FNN, although our TSE-Net contains much fewer parameters, it achieves better or comparable classification results in all kinds of tasks. Our method is also shown to learn models with better interpretability, which is also an important problem in deep learning. In the future, we will generalize our method to other networks like RNNs and CNNs.

\begin{small}
\bibliographystyle{named}
\bibliography{TSE-Net}
\end{small}

\end{document}